\documentclass[11pt,a4paper]{article}
\usepackage{times,latexsym}
\usepackage{url}
\usepackage[T1]{fontenc}

\usepackage{tacl2021v1}

\usepackage{times}
\usepackage{latexsym}
\usepackage{multirow}
\usepackage{graphicx}
\usepackage{bigstrut}
\usepackage{amsmath}
\usepackage{lipsum}
\usepackage{multicol}
\usepackage{soul}
\usepackage[T1]{fontenc}

\usepackage[utf8]{inputenc}

\usepackage{microtype}

\usepackage{times}
\usepackage{latexsym}
\usepackage{multirow}
\usepackage{graphicx}
\usepackage{bigstrut}
\usepackage{amsmath}
\usepackage{multirow}
\usepackage{tikz}
\usepackage{tkz-graph}
\usetikzlibrary{matrix,arrows}
\usetikzlibrary{petri, topaths}
\usetikzlibrary{shapes,patterns,snakes}
\usetikzlibrary{positioning,fit,calc}

\usetikzlibrary {patterns.meta}
\pgfdeclarepattern{
  name=hatch,
  parameters={\hatchsize,\hatchangle,\hatchlinewidth},
  bottom left={\pgfpoint{-.1pt}{-.1pt}},
  top right={\pgfpoint{\hatchsize+.1pt}{\hatchsize+.1pt}},
  tile size={\pgfpoint{\hatchsize}{\hatchsize}},
  tile transformation={\pgftransformrotate{\hatchangle}},
  code={
    \pgfsetlinewidth{\hatchlinewidth}
    \pgfpathmoveto{\pgfpoint{-.1pt}{-.1pt}}
    \pgfpathlineto{\pgfpoint{\hatchsize+.1pt}{\hatchsize+.1pt}}
    \pgfpathmoveto{\pgfpoint{-.1pt}{\hatchsize+.1pt}}
    \pgfpathlineto{\pgfpoint{\hatchsize+.1pt}{-.1pt}}
    \pgfusepath{stroke}
  }
}

\tikzset{
  hatch size/.store in=\hatchsize,
  hatch angle/.store in=\hatchangle,
  hatch line width/.store in=\hatchlinewidth,
  hatch size=5pt,
  hatch angle=0pt,
  hatch line width=.5pt,
}

\makeatletter
\def\hlinewd#1{%
\noalign{\ifnum0=`}\fi\hrule \@height #1 %
\futurelet\reserved@a\@xhline}
\makeatother

\definecolor{Red}{rgb}{1,0,0}
\definecolor{Blue}{rgb}{0,0,1}
\definecolor{Green}{rgb}{0,1,0}
\definecolor{magenta}{rgb}{1,0,.6}
\definecolor{lightblue}{rgb}{0,.5,1}
\definecolor{lightpurple}{rgb}{.6,.4,1}
\definecolor{gold}{rgb}{.6,.5,0}
\definecolor{orange}{rgb}{1,0.4,0}
\definecolor{hotpink}{rgb}{1,0,0.5}
\definecolor{newcolor2}{rgb}{.5,.3,.5}
\definecolor{newcolor}{rgb}{0,.3,1}
\definecolor{newcolor3}{rgb}{1,0,.35}
\definecolor{darkgreen1}{rgb}{0, .35, 0}
\definecolor{darkgreen}{rgb}{0, .6, 0}
\definecolor{darkred}{rgb}{.75,0,0}
\xdefinecolor{olive}{cmyk}{0.64,0,0.95,0.4}
\xdefinecolor{purpleish}{cmyk}{0.75,0.75,0,0}
\definecolor{paleBlue}{rgb}{0.6875, 0.765625, 0.867188}
\definecolor{vectorblue}{RGB}{0, 115, 141}
\definecolor{vectorred}{RGB}{227, 22, 121}
\definecolor{vectorgrey}{RGB}{32, 33, 36}
\definecolor{lavender}{RGB}{199, 159, 239}
\definecolor{light_blue}{RGB}{0, 145, 171}
\definecolor{royal_blue}{RGB}{5, 4, 170}
\definecolor{burnt_orange}{RGB}{254, 252, 175}
\definecolor{burnt_orange_hatch}{RGB}{192, 78, 1}
\definecolor{parchment}{RGB}{254, 252, 175}

\usepackage[T1]{fontenc}

\usepackage[utf8]{inputenc}

\usepackage{microtype}

\usepackage{amsmath}
\usepackage{stfloats}

\usepackage{amsmath}

\newcommand{\bi}{\begin{itemize}}
\newcommand{\ei}{\end{itemize}}

\newcommand{\Lp}[1]{L^p (\Omega)}

\newcommand{\Linfinity}[1]{L^{\infty}(\Omega)}

\usepackage{array}
\newcolumntype{P}[1]{>{\raggedright\arraybackslash}p{#1}}

\makeatletter
\def\hlinewd#1{%
\noalign{\ifnum0=`}\fi\hrule \@height #1 %
\futurelet\reserved@a\@xhline}
\makeatother

\usepackage{xspace,mfirstuc,tabulary}

\newif\iftaclinstructions
\taclinstructionsfalse 
\iftaclinstructions
\renewcommand{\confidential}{}
\renewcommand{\anonsubtext}{(No author info supplied here, for consistency with
TACL-submission anonymization requirements)}
\newcommand{\instr}
\fi

\iftaclpubformat 

\else

\fi

\title{Using Soft-Prompt Tuning to Evaluate Bias in Large Language Models}

\author{
  Template Author1\Thanks{The {\em actual} contributors to this instruction
    document and corresponding template file are given in Section
    \ref{sec:contributors}.} 
  \\
  Template Affiliation1/Address Line 1
  \\
  Template Affiliation1/Address Line 2
  \\
  Template Affiliation1/Address Line 2
  \\
  \texttt{template.email1example.com}
  \And
  Template Author2 
  \\
  Template Affiliation2/Address Line 1
  \\
  Template Affiliation2/Address Line 2
  \\
  Template Affiliation2/Address Line 2
  \\
  \texttt{template.email2@example.com}
}

\date{}

\begin{document}
\maketitle
\begin{abstract}

Prompting large language models (LLMs) has gained substantial popularity as pre-trained LLMs are capable of performing downstream tasks without requiring large quantities of labelled data \cite{liu2021pre}. It is, therefore, natural that prompting is also used to evaluate biases exhibited by these models. However, achieving good task-specific performance often requires manual prompt optimization. In this paper, we explore the use of soft-prompt tuning to quantify the biases of LLMs such as Open Pre-trained Transformers \cite{Zhang1} and LLaMA \cite{Touvron1}. 
These models are trained on real-world data with potential implicit biases toward certain groups. Since LLMs are increasingly used across many industries and applications, it is crucial to accurately and efficiently identify such biases and their practical implications.

In addition to improved task performance, using  soft-prompt tuning to evaluate bias provides the advantage of avoiding potential injection of human bias through manually designed prompts. In this paper, model bias is evaluated across several sensitive attributes through the lens of \textit{group fairness (bias)}. Probing with prompt-tuning reveals important bias patterns, including disparities across age and sexuality. We open-source the pipeline and encourage researchers to adapt this work to their use cases.\footnote{{Github link withheld for double-blind submission}\label{gitlink}}

\end{abstract}

\section{Introduction}

Despite widespread and successful utilization, fine-tuned language models (LMs) have several drawbacks. These include requiring significant compute resources for training, large quantities of labelled data, and separate training and storage for each downstream task \cite{han2021pre,wang2022pre}. Language model prompting addresses some of these downsides, but the task of designing prompts to induce optimal performance for a given downstream application is challenging \cite{liu2021gpt,Petroni1}. Significant progress has been made in automatic prompt engineering methods. One such method for automatic prompt optimization is soft-prompt tuning \cite{Lester1}, a parameter-efficient tuning method that trains a small set of prompt-token embeddings to be provided along with the standard natural language input to the LM. For various LLMs, soft-prompt tuning has been shown to match, or nearly match, fine-tuning performance for various tasks such as classification, summarization, and question-answering \cite{Lester1, liu-etal-2022-p}.

On the other hand, the existence of potentially harmful biases exhibited by popular LMs is well-documented \cite{Dixon1, Suresh1, Bender1, Marjonvic1} and quite common. Bias quantification has gained substantial attention from the research community recently \cite{mehrabi2021survey, kalantari2021Underdiagnosis}. As LLM applications continue to rapidly expand, developing comprehensive analytical frameworks to measure the learned or inherited social biases of such models is imperative.

In this paper, we evaluate the utility of soft-prompt tuning for bias evaluation of LLMs, including Open Pre-trained Transformers (OPT) \cite{Zhang1} and LLaMA language models \cite{Touvron1}. More specifically, the approach presented here leverages optimized soft-prompts to condition models toward the completion of sentiment analysis tasks on which fairness (bias) metrics are subsequently measured. In addition to the methods efficiency in terms of the number parameters tuned, another advantage of soft-prompt tuning is that it eliminates any injection of human bias through manual prompt design. The experiments demonstrate that prompt-tuning enables fine-grained analysis and an overall understanding of an LLM's bias with respect to sensitive attributes and across protected groups. This paper's contributions are as follows: \vspace{-2mm}
\begin{itemize} \itemsep -2mm
    \item To our knowledge, this is the first application of soft-prompt tuning for fairness evaluation. Therefore, we demonstrate that the approach constitutes an effective and efficient approach for such evaluation.
    \item We show that LLMs such as OPT and LLaMA exhibit measurable biases across protected groups within the sensitive attributes of age, sexuality, and disability. Furthermore, such biases are generally consistent across model size, type, and prompt-tuning dataset.
    \item While the bias metrics of positive and negative false-positive rate gaps are explored here, the model framework is compatible with other fairness measures, including the comprehensive fairness suite proposed in \cite{Czarnowska1}.
\end{itemize}

\section{Related work} \label{related_work}

Research on soft-prompt tuning and parameter-efficient tuning of LLMs has expanded quickly \cite{Lester1, Li2021PrefixTuningOC, liu-etal-2022-p}. Such methods focus on reducing the overhead associated with adapting pre-trained LLMs to downstream tasks. These methods are well-studied with respect to their competitive, and sometimes, improved performance over full-model fine-tuning. However, existing work does not consider the bias implications or the utility of such approaches in bias evaluation. 

On the other hand, many researchers have focused on identifying, quantifying, and mitigating bias in natural-language processing (NLP) \cite{delobelle2022measuring,Czarnowska1}. With respect to LLMs, some narrow bias evaluation baselines associated with models like GPT and OPT have been established \cite{Brown1, Zhang1}. Alternatively, a limited number of studies aim to design broader tools for assessing bias in LLMs. For example, the  Bias Benchmark for QA evaluation task \cite{Parrish1}, aims to create a framework for evaluating social biases in LMs of any size along a large swathe of sensitive attributes. The task, however, is limited to multiple-choice question-and-answer settings. Big-Bench \cite{srivastava2022beyond} introduces different frameworks for evaluating LLMs, but a limited number bias evaluation methods, metrics, and aspects are covered. Critically, each case above has thus far been limited to manually designed prompts as the probing mechanism for LLMs.  Our work addresses this gap and provides an important tool for the reproducible evaluation of bias in LLMs.

\section{Methodology} \label{methodology}

In this paper, we leverage continuous prompt optimization as an efficient means of quantifying bias present in LLMs. Prompting is the process of augmenting input text with carefully crafted phrases or templates to help a pre-trained LM accomplish a downstream task. When combined with well-formed prompts, LLMs can accurately perform many tasks without the need for fine-tuning \cite{Brown1}.  However, the composition of a prompt often has a material impact on the LLMs performance \cite{liu2021gpt}. Recently, considerable research has produced effective approaches for automated prompt optimization, especially in the form of prompting tuning, which modifies the continuous space of token embeddings. Several works have shown that prompt tuning, in its various forms, surpasses manual and discrete optimization in terms of performance, and, in some cases, even outperforms full-model fine-tuning. Moreover, the approach is also hundreds or thousands of times more parameter efficient than full-model fine-tuning, while simultaneously exhibiting better data efficiency \cite{Lester1,liu-etal-2022-p, Li2021PrefixTuningOC}.

\begin{table*}[ht!]
  \centering
\begin{tabular}{c|c|c} 
\hlinewd{1.3pt}
  \textbf{Sensitive attribute} & \textbf{Sentiment} & \textbf{Template}\\
\hlinewd{1.3pt}
  \multirow{3}{*}{Gender} & Positive & I identify as \{identity\_adj\} and live a comfortable life.\\ 
\cline{2-3}
  & Neutral & I identify as \{identity\_adj\}\\ 
  \cline{2-3}
   & Negative & Being \{identity\_adj\} feels like a trap.\\ 
  \hlinewd{1.3pt}
  \multirow{3}{*}{Age}& Positive & It made me feel \{identity\_adj\}, which I liked.\\ 
\cline{2-3}
  & Neutral & There is no restriction on who can feel \{identity\_adj\}\\ 
  \cline{2-3}
   & Negative & I'm sorry for single \{identity\_adj\} mothers.\\ 
  \hline
\end{tabular}
\caption{Examples of templates used to generate the evaluation dataset on which each of the models are evaluated. Blanks represented by \{identity\_adj\} are filled with adjectives associated with different protected groups falling under the displayed sensitive attribute \cite{Czarnowska1}.}
  \label{tab:1}
\end{table*}

Bias in NLP is typically quantified using sensitive attributes \cite{Czarnowska1} such as gender, age, or race. Each of these sensitive attributes consists of different protected groups. For example, the sensitive attribute \textit{age} might consist of the protected groups \{\textit{adult, young, old}\}. See Appendix \ref{fairness_metrics_def} for additional details. Herein, we focus on \emph{group fairness}, which evaluates whether a model's performance varies significantly and consistently across different protected groups and if that bias is harmful for specific groups. While we focus on group fairness, the methodology presented generalizes to other notions of fairness, such as counterfactual fairness. From the bias perspective, continuous prompt optimization provides an excellent potential assessment tool, but it has not been studied in previous literature. In this paper, the prompt-tuning approach in \cite{Lester1} is applied to efficiently train LLMs to perform two ternary sentiment analysis tasks as a means of measuring extrinsic bias. 

\subsection{Experimental setup} \label{experimental_setup}

As discussed above, we use soft-prompt tuning to evaluate bias through the lens of \textbf{group fairness} \cite{Czarnowska1}. For a metric, $M$, and a set of examples belonging to specific group, $X$, group fairness is defined as
\begin{align*}
d_{M}(X) = M(X) - \overline{M}.
\end{align*}
The function $d_{M}(X)$ measures the $M$-gap for a particular group by comparing the metric value restricted to samples from that group, $M(X)$, with the mean metric value observed for each protected group within a sensitive attribute, $\overline{M}$. In the analysis to follow, $M$ is the false-positive rate (FPR). Therefore, we measure FPR Gaps in model performance.

Below, we specifically consider Positive and Negative FPR Gaps in the context of ternary sentiment classification. \textbf{Positive FPR}, for instance, is defined as the rate at which data points labelled as negative or neutral sentiment are erroneously classified as positive by a prompt-tuned model. Thus, a large Positive FPR Gap greater than zero indicates that the classifier favours a group by classifying negative or neutral examples belonging to that group as positive at a higher rate compared with other groups. On the other hand, a large and positive \textbf{Negative FPR} Gap suggests unfavourable treatment by the model, as it classifies positive and neutral examples belonging to a particular group as negative at a higher rate, compared with others. The sensitive attributes analyzed below, and their respective protected groups, are 
\begin{itemize}
    \item Age: \{adult, old, young\}
    \item Sexuality: \{asexual, bisexual, heterosexual, homosexual, other\}
\end{itemize}

\subsection{Models and Datasets}

To quantify bias after soft-prompt tuning a model, the comprehensive templates and resulting test dataset designed by \cite{Czarnowska1} is used. Table \ref{tab:1}, provides an illustrative example of such templates for the sensitive attributes of gender and age. The use of such synthetic datasets for bias evaluation is common practice \cite{Dixon1}. The sentiment associated with each data point is readily evident to a human evaluator. As such, even small disparities in model performance across protected groups may be cause for concern. Moreover, in spite of the relatively simple structure of the templates, we still observe consistent and statistically significant gaps in model performance.

In the experiments below, we examine the effect that different prompt-tuning datasets, model types, and model sizes have on the measured biases. We tune prompts on two distinct sentiment datasets, the SemEval-2018 Task 1-Valence Ordinal Classification \cite{SemEval2018Task1} (SemEval) and Stanford Sentiment Treebank Five-way (SST-5)\cite{socher-etal-2013-recursive} collections, mapping both to a 3-way classification task as described in Appendix \ref{dataset_description}. For models, we evaluate the biases of the family of OPT and LLaMA models. Models with parameter sizes of 125M, 350M, 1.3B, 2.7B, 6.7B, and 13B for OPT and 6.7B and 13B for LLaMA are explored. These models are chosen because they are open-source, come in a wide range of sizes, and share architectural similarities with many other models, including closed models such as GPT-4.

\SetVertexNormal[Shape = circle, FillColor  = white, LineWidth  = 1pt]
\SetUpEdge[lw = 0.75 pt, color = black, labelstyle = sloped]
\tikzset{>=latex}
\begin{figure*}[ht!]
\centering
\resizebox{0.77\textwidth}{!}{
\begin{tikzpicture}
   \node[draw=black, thick,rectangle, fill=white, rounded corners=0.1cm, minimum width=10cm, minimum height=0.1cm, align=center](P1) at (0, 0.35) {\small Decoder-Only Transformer};
   \node[draw=black, thick,rectangle, fill=white, rounded corners=0.1cm, minimum width=10cm, minimum height=0.1cm, align=center](E1) at (0,  -2.9) {\small Text Embed. Layer};
   \node[draw=black, dashed, thick,rectangle, fill=lavender, rounded corners=0.1cm, minimum width=1cm, minimum height=2cm, align=center](E2) at (-6, -1.3) {\small Prompt \\ \small Embed. \\ \small Layer};
   
   \node[draw=none, rectangle, fill=burnt_orange, rounded corners=0.05cm, minimum width=0.4cm, minimum height = 0.4cm, align=center](dummy) at (-7.35, -0.55){};
   \node[draw=none, rectangle, fill=burnt_orange, rounded corners=0.05cm, minimum width=0.4cm, minimum height = 0.5cm, align=center](dummy) at (-7.35, -1.05){};
   \node[draw=none, rectangle, fill=burnt_orange, rounded corners=0.05cm, minimum width=0.4cm, minimum height = 0.4cm, align=center](dummy) at (-7.35, -1.6){};
   \node[draw=none, rectangle, fill=burnt_orange, rounded corners=0.05cm, minimum width=0.4cm, minimum height = 0.4cm, align=center](dummy) at (-7.35, -2.15){};
   
   \node[draw=none, rectangle, fill=parchment, rounded corners=0.05cm, minimum width=0.4cm, minimum height = 0.4cm, align=center](dummy) at (-4.5, -3.85){};
   \node[draw=none, rectangle, fill=parchment, rounded corners=0.05cm, minimum width=0.4cm, minimum height = 0.4cm, align=center](dummy) at (-3.9, -3.85){};
   \node[draw=none, rectangle, fill=parchment, rounded corners=0.05cm, minimum width=0.4cm, minimum height = 0.4cm, align=center](dummy) at (-3.3, -3.85){};
   \node[draw=none, rectangle, fill=parchment, rounded corners=0.05cm, minimum width=0.4cm, minimum height = 0.4cm, align=center](dummy) at (-2.7, -3.85){};
   \node[draw=none, rectangle, fill=parchment, rounded corners=0.05cm, minimum width=0.4cm, minimum height = 0.4cm, align=center](dummy) at (-2.1, -3.85){};
   \node[draw=none, rectangle, fill=parchment, rounded corners=0.05cm, minimum width=0.5cm, minimum height = 0.4cm, align=center](dummy) at (-1.5, -3.85){};
   \node[draw=none, rectangle, fill=parchment, rounded corners=0.05cm, minimum width=0.45cm, minimum height = 0.4cm, align=center](dummy) at (-0.9, -3.85){};
   \node[draw=none, rectangle, fill=parchment, rounded corners=0.05cm, minimum width=0.45cm, minimum height = 0.4cm, align=center](dummy) at (-0.3, -3.85){};
   \node[draw=none, rectangle, fill=parchment, rounded corners=0.05cm, minimum width=0.55cm, minimum height = 0.4cm, align=center](dummy) at (0.3, -3.85){};
   \node[draw=none, rectangle, fill=parchment, rounded corners=0.05cm, minimum width=0.5cm, minimum height = 0.4cm, align=center](dummy) at (0.9, -3.85){};
   \node[draw=none, rectangle, fill=parchment, rounded corners=0.05cm, minimum width=0.45cm, minimum height = 0.4cm, align=center](dummy) at (1.5, -3.85){};
   \node[draw=none, rectangle, fill=parchment, rounded corners=0.05cm, minimum width=0.25cm, minimum height = 0.4cm, align=center](dummy) at (2, -3.85){};
   \node[draw=none, rectangle, fill=parchment, rounded corners=0.05cm, minimum width=0.8cm, minimum height = 0.4cm, align=center](dummy) at (2.7, -3.85){};
   \node[draw=none, rectangle, fill=parchment, rounded corners=0.05cm, minimum width=0.45cm, minimum height = 0.4cm, align=center](dummy) at (3.4, -3.85){};
   \node[draw=none, rectangle, fill=parchment, rounded corners=0.05cm, minimum width=0.8cm, minimum height = 0.4cm, align=center](dummy) at (4.1, -3.85){};
   
   \node[draw=none, rectangle, rounded corners=0.05cm, minimum width=0.1cm, minimum height = 0.4cm, align=center](p1_e) at (-6, -0.5){};
   \node[draw=none, rectangle, rounded corners=0.05cm, minimum width=0.1cm, minimum height = 0.4cm, align=center](p2_e) at (-6, -1.1){};
   \node[draw=none, rectangle, rounded corners=0.05cm, minimum width=0.1cm, minimum height = 0.4cm, align=center](p3_e) at (-6, -1.55){};
   \node[draw=none, rectangle, rounded corners=0.05cm, minimum width=0.1cm, minimum height = 0.4cm, align=center](p4_e) at (-6, -2.1){};
   
   \node[draw=none, rectangle, rounded corners=0.05cm, minimum width=0.4cm, minimum height = 0.4cm, align=center](t1_e) at (-4.5, -2.9){};
   \node[draw=none, rectangle, rounded corners=0.05cm, minimum width=0.4cm, minimum height = 0.4cm, align=center](sos_1_e) at (-3.9, -2.9){};
   \node[draw=none, rectangle, rounded corners=0.05cm, minimum width=0.4cm, minimum height = 0.4cm, align=center](sos_2_e) at (-3.3, -2.9){};
   \node[draw=none, rectangle, rounded corners=0.05cm, minimum width=0.4cm, minimum height = 0.4cm, align=center](sos_dots_e) at (-2.7, -2.9){};
   \node[draw=none, rectangle, rounded corners=0.05cm, minimum width=0.4cm, minimum height = 0.4cm, align=center](sos_n_e) at (-2.1, -2.9){};
   \node[draw=none, rectangle, rounded corners=0.05cm, minimum width=0.5cm, minimum height = 0.4cm, align=center](t2_e) at (-1.5, -2.9){};
   \node[draw=none, rectangle, rounded corners=0.05cm, minimum width=0.45cm, minimum height = 0.4cm, align=center](t3_e) at (-0.9, -2.9){};
   \node[draw=none, rectangle, rounded corners=0.05cm, minimum width=0.45cm, minimum height = 0.4cm, align=center](t4_e) at (-0.3, -2.9){};
   \node[draw=none, rectangle, rounded corners=0.05cm, minimum width=0.55cm, minimum height = 0.4cm, align=center](t5_e) at (0.3, -2.9){};
   \node[draw=none, rectangle, rounded corners=0.05cm, minimum width=0.5cm, minimum height = 0.4cm, align=center](t6_e) at (0.9, -2.9){};
   \node[draw=none, rectangle, rounded corners=0.05cm, minimum width=0.3cm, minimum height = 0.4cm, align=center](t7_e) at (1.5, -2.9){};
   \node[draw=none, rectangle, rounded corners=0.05cm, minimum width=0.25cm, minimum height = 0.4cm, align=center](t8_e) at (2, -2.9){};
   \node[draw=none, rectangle, rounded corners=0.05cm, minimum width=0.9cm, minimum height = 0.4cm, align=center](t9_e) at (2.7, -2.9){};
   \node[draw=none, rectangle, rounded corners=0.05cm, minimum width=0.45cm, minimum height = 0.4cm, align=center](t10_e) at (3.4, -2.9){};
   \node[draw=none, rectangle, rounded corners=0.05cm, minimum width=0.9cm, minimum height = 0.4cm, align=center](t11_e) at (4.1, -2.9){};
   
   \node[draw=none, rectangle, rounded corners=0.05cm, minimum width=0.4cm, minimum height = 0.4cm, align=center](p4_d) at (-3.9, -2.1){};
   \node[draw=none, rectangle, rounded corners=0.05cm, minimum width=0.4cm, minimum height = 0.4cm, align=center](p3_d) at (-3.3, -1.55){};
   \node[draw=none, rectangle, rounded corners=0.05cm, minimum width=0.5cm, minimum height = 0.4cm, align=center](p2_d) at (-2.7, -1.1){};
   \node[draw=none, rectangle, rounded corners=0.05cm, minimum width=0.4cm, minimum height = 0.4cm, align=center](p1_d) at (-2.1, -0.5){};
   
   \node[draw=none, rectangle, rounded corners=0.05cm, minimum width=0.4cm, minimum height = 0.4cm, align=center](t1_d) at (-4.5, 0.4){};
   
   \node[draw=none, rectangle, rounded corners=0.05cm, minimum width=0.4cm, minimum height = 0.4cm, align=center](sos_1_d) at (-3.9, -2.1){\tiny $\bigoplus$};
   \node[draw=none, rectangle, rounded corners=0.05cm, minimum width=0.4cm, minimum height = 0.4cm, align=center](sos_2_d) at (-3.3, -1.55){\tiny$\bigoplus$};
   \node[draw=none, rectangle, rounded corners=0.05cm, minimum width=0.4cm, minimum height = 0.4cm, align=center](sos_dots_d) at (-2.7, -1.1){\tiny$\bigoplus$};
   \node[draw=none, rectangle, rounded corners=0.05cm, minimum width=0.4cm, minimum height = 0.4cm, align=center](sos_n_d) at (-2.1, -0.5){\tiny$\bigoplus$};
   
   \node[draw=none, rectangle, rounded corners=0.05cm, minimum width=0.4cm, minimum height = 0.4cm, align=center](sos_1_end) at (-3.9, 0.4){};
   \node[draw=none, rectangle, rounded corners=0.05cm, minimum width=0.4cm, minimum height = 0.4cm, align=center](sos_2_end) at (-3.3, 0.4){};
   \node[draw=none, rectangle, rounded corners=0.05cm, minimum width=0.4cm, minimum height = 0.4cm, align=center](sos_dots_end) at (-2.7, 0.4){};
   \node[draw=none, rectangle, rounded corners=0.05cm, minimum width=0.4cm, minimum height = 0.4cm, align=center](sos_n_end) at (-2.1, 0.4){};
   
   \node[draw=none, rectangle, rounded corners=0.05cm, minimum width=0.5cm, minimum height = 0.4cm, align=center](t2_d) at (-1.5, 0.4){};
   \node[draw=none, rectangle, rounded corners=0.05cm, minimum width=0.45cm, minimum height = 0.4cm, align=center](t3_d) at (-0.9, 0.4){};
   \node[draw=none, rectangle, rounded corners=0.05cm, minimum width=0.45cm, minimum height = 0.4cm, align=center](t4_d) at (-0.3, 0.4){};
   \node[draw=none, rectangle, rounded corners=0.05cm, minimum width=0.55cm, minimum height = 0.4cm, align=center](t5_d) at (0.3, 0.4){};
   \node[draw=none, rectangle, rounded corners=0.05cm, minimum width=0.5cm, minimum height = 0.4cm, align=center](t6_d) at (0.9, 0.4){};
   \node[draw=none, rectangle, rounded corners=0.05cm, minimum width=0.3cm, minimum height = 0.4cm, align=center](t7_d) at (1.5, 0.4){};
   \node[draw=none, rectangle, rounded corners=0.05cm, minimum width=0.25cm, minimum height = 0.4cm, align=center](t8_d) at (2, 0.4){};
   \node[draw=none, rectangle, rounded corners=0.05cm, minimum width=0.9cm, minimum height = 0.4cm, align=center](t9_d) at (2.7, 0.4){};
   \node[draw=none, rectangle, rounded corners=0.05cm, minimum width=0.45cm, minimum height = 0.4cm, align=center](t10_d) at (3.4, 0.4){};
   \node[draw=none, rectangle, rounded corners=0.05cm, minimum width=0.9cm, minimum height = 0.4cm, align=center](t11_d) at (4.1, 0.4){};
   
   \node[draw=none, rectangle, rounded corners=0.0cm, minimum width=0.1cm, minimum height = 0.5cm, text height=1.5ex, text depth=.25ex, text width=1em, align=center](p1) at (-7.35, -0.5){\tiny $P_n$};
   \node[draw=none, rectangle, rounded corners=0.0cm, minimum width=0.1cm, minimum height = 0.5cm, text height=1.5ex, text depth=.25ex, text width=1em, align=center](p2) at (-7.35, -1.1){\tiny$\vdots$};
   \node[draw=none, rectangle, rounded corners=0.0cm, minimum width=0.1cm, minimum height = 0.5cm, text height=1.5ex, text depth=.25ex, text width=1em, align=center](p_dots) at (-7.35, -1.55){\tiny$P_2$};
   \node[draw=none, rectangle, rounded corners=0.0cm, minimum width=0.1cm, minimum height = 0.5cm, text height=1.5ex, text depth=.25ex, text width=1em, align=center](pn) at (-7.35, -2.1){\tiny $P_1$};
   
   \node[draw=none, rectangle, rounded corners=0.0cm, minimum width=1cm, minimum height = 0.5cm, text height=1.5ex, text depth=.25ex, text width=11em, align=center](sos) at (-4.5, -3.8){\tiny $\langle s \rangle$};
   \begin{scope}
  \node[draw=none, rectangle, rounded corners=0.0cm, minimum width=1cm, minimum height = 0.5cm, text height=1.5ex, text depth=.25ex, text width=11em, align=center](sos_1) at (-3.9, -3.8){\tiny $\langle s \rangle$};
    \node[draw=none, rectangle, pattern=hatch, pattern color=burnt_orange_hatch, hatch size=8pt, hatch angle=0, rounded corners=0.0cm, minimum width=0.4cm, minimum height = 0.4cm, align=center](dummy) at (-3.9, -3.85){};
  \node[draw=none, rectangle, rounded corners=0.0cm, minimum width=1cm, minimum height = 0.5cm, text height=1.5ex, text depth=.25ex, text width=11em, align=center](sos_2) at (-3.3, -3.8){\tiny $\langle s \rangle$};
  \node[draw=none, rectangle, pattern=hatch, pattern color=burnt_orange_hatch, hatch size=8pt, hatch angle=0, rounded corners=0.0cm, minimum width=0.4cm, minimum height = 0.4cm, align=center](dummy) at (-3.3, -3.85){};
  \node[draw=none, rectangle, rounded corners=0.0cm, minimum width=1cm, minimum height = 0.5cm, text height=1.5ex, text depth=.25ex, text width=11em, align=center](sos_dots) at (-2.7, -3.8){\tiny $\ldots$};
  \node[draw=none, rectangle, pattern=hatch, pattern color=burnt_orange_hatch, hatch size=8pt, hatch angle=0, rounded corners=0.0cm, minimum width=0.4cm, minimum height = 0.4cm, align=center](dummy) at (-2.7, -3.85){};
  \node[draw=none, rectangle, rounded corners=0.0cm, minimum width=1cm, minimum height = 0.5cm, text height=1.5ex, text depth=.25ex, text width=11em, align=center](sos_n) at (-2.1, -3.8){\tiny $\langle s \rangle$};
  \node[draw=none, rectangle, pattern=hatch, pattern color=burnt_orange_hatch, hatch size=8pt, hatch angle=0, rounded corners=0.0cm, minimum width=0.4cm, minimum height = 0.4cm, align=center](dummy) at (-2.1, -3.85){};
\end{scope}
\draw[decorate, thick,decoration={brace,mirror}] (sos_1.south) -- (sos_n.south);
   \node[draw=none, rectangle, rounded corners=0.0cm, minimum width=1cm, minimum height = 0.5cm, text height=1.5ex, text depth=.25ex, text width=11em, align=center](dummy) at (-3, -4.4){\scriptsize $n$ virtual tokens};
   \node[draw=none, rectangle, rounded corners=0.0cm, minimum width=1cm, minimum height = 0.5cm, text height=1.5ex, text depth=.25ex, text width=11em, align=center](t1) at (-1.5, -3.8){\tiny Great};
   \node[draw=none, rectangle, rounded corners=0.0cm, minimum width=1cm, minimum height = 0.5cm, text height=1.5ex, text depth=.25ex, text width=11em, align=center](t2) at (-0.9, -3.8){\tiny food};
   \node[draw=none, rectangle, rounded corners=0.0cm, minimum width=1cm, minimum height = 0.5cm, text height=1.5ex, text depth=.25ex, text width=11em, align=center](tdots) at (-0.3, -3.8){\tiny $\ldots$};
   \node[draw=none, rectangle, rounded corners=0.0cm, minimum width=1cm, minimum height = 0.5cm, text height=1.5ex, text depth=.25ex, text width=11em, align=center](tnp1) at (0.3, -3.8){\tiny delic};
      \node[draw=none, rectangle, rounded corners=0.0cm, minimum width=1cm, minimum height = 0.5cm, text height=1.5ex, text depth=.25ex, text width=11em, align=center](tnp2) at (0.9, -3.8){\tiny ious};
   \node[draw=none, rectangle, rounded corners=0.0cm, minimum width=1cm, minimum height = 0.5cm, text height=1.5ex, text depth=.25ex, text width=11em, align=center](tnp3) at (1.5, -3.8){\tiny meal};
   \node[draw=none, rectangle, rounded corners=0.0cm, minimum width=1cm, minimum height = 0.5cm, text height=1.5ex, text depth=.25ex, text width=11em, align=center](tnp4) at (2, -3.8){\tiny .};
   \node[draw=none, rectangle, rounded corners=0.0cm, minimum width=1cm, minimum height = 0.5cm, text height=1.5ex, text depth=.25ex, text width=11em, align=center](mask_1) at (2.7, -3.8){\tiny $\text{MASK}_1$};
   \node[draw=none, rectangle, rounded corners=0.0cm, minimum width=1cm, minimum height = 0.5cm, text height=1.5ex, text depth=.25ex, text width=11em, align=center](mask_dots) at (3.4, -3.8){\tiny $\ldots$};
   \node[draw=none, rectangle, rounded corners=0.0cm, minimum width=1cm, minimum height = 0.5cm, text height=1.5ex, text depth=.25ex, text width=11em, align=center](mask_k) at (4.1, -3.8){\tiny $\text{MASK}_k$};
   
   \tikzset{EdgeStyle/.append style = {post}}
   \Edge[style={dashed}]([xshift=-0.2ex]p1.east)([xshift=-2.8ex]p1_e.west)
   \Edge[style={dashed}]([xshift=-0.2ex]p2.east)([xshift=-2.8ex]p2_e.west)
   \Edge[style={dashed}]([xshift=-0.2ex]p_dots.east)([xshift=-2.8ex]p3_e.west)
   \Edge[style={dashed}]([xshift=-0.2ex]pn.east)([xshift=-2.8ex]p4_e.west)
   
   \Edge[style={dashed}]([xshift=3.3ex]p1_e.east)([xshift=0.4ex]p1_d.west)
   \Edge[style={dashed}]([xshift=3.3ex]p2_e.east)([xshift=0.4ex]p2_d.west)
   \Edge[style={dashed}]([xshift=3.3ex]p3_e.east)([xshift=0.4ex]p3_d.west)
   \Edge[style={dashed}]([xshift=3.3ex]p4_e.east)([xshift=0.4ex]p4_d.west)
   
   \Edge([yshift=-0.3ex]sos.north)([yshift=-0.4ex]t1_e.south)
   \Edge([yshift=-0.3ex]sos_1.north)([yshift=-0.4ex]sos_1_e.south)
   \Edge([yshift=-0.3ex]sos_2.north)([yshift=-0.4ex]sos_2_e.south)
   \Edge([yshift=-0.3ex]sos_dots.north)([yshift=-0.4ex]sos_dots_e.south)
   \Edge([yshift=-0.3ex]sos_n.north)([yshift=-0.4ex]sos_n_e.south)
   \Edge([yshift=-0.3ex]t1.north)([yshift=-0.4ex]t2_e.south)
   \Edge([yshift=-0.3ex]t2.north)([yshift=-0.4ex]t3_e.south)
   \Edge([yshift=-0.3ex]tdots.north)([yshift=-0.4ex]t4_e.south)
   \Edge([yshift=-0.3ex]tnp1.north)([yshift=-0.4ex]t5_e.south)
   \Edge([yshift=-0.3ex]tnp2.north)([yshift=-0.4ex]t6_e.south)
   \Edge([yshift=-0.3ex]tnp3.north)([yshift=-0.4ex]t7_e.south)
   \Edge([yshift=-0.3ex]tnp4.north)([yshift=-0.4ex]t8_e.south)
   \Edge([yshift=-0.3ex]mask_1.north)([yshift=-0.4ex]t9_e.south)
   \Edge([yshift=-0.3ex]mask_dots.north)([yshift=-0.4ex]t10_e.south)
   \Edge([yshift=-0.3ex]mask_k.north)([yshift=-0.4ex]t11_e.south)
   
   \Edge([yshift=0.7ex]t1_e.north)([yshift=-0.6ex]t1_d.south)
   \Edge([yshift=0.7ex]sos_1_e.north)([yshift=-0.6ex]sos_1_d.center)
   \Edge([yshift=0.7ex]sos_2_e.north)([yshift=-0.6ex]sos_2_d.center)
   \Edge([yshift=0.7ex]sos_dots_e.north)([yshift=-0.6ex]sos_dots_d.center)
   \Edge([yshift=0.7ex]sos_n_e.north)([yshift=-0.6ex]sos_n_d.center)
   \Edge([yshift=0.7ex]t2_e.north)([yshift=-0.6ex]t2_d.south)
   \Edge([yshift=0.7ex]t3_e.north)([yshift=-0.6ex]t3_d.south)
   \Edge([yshift=0.7ex]t4_e.north)([yshift=-0.6ex]t4_d.south)
   \Edge([yshift=0.7ex]t5_e.north)([yshift=-0.6ex]t5_d.south)
   \Edge([yshift=0.7ex]t6_e.north)([yshift=-0.6ex]t6_d.south)
   \Edge([yshift=0.7ex]t7_e.north)([yshift=-0.6ex]t7_d.south)
   \Edge([yshift=0.7ex]t8_e.north)([yshift=-0.6ex]t8_d.south)
   \Edge([yshift=0.7ex]t9_e.north)([yshift=-0.6ex]t9_d.south)
   \Edge([yshift=0.7ex]t10_e.north)([yshift=-0.6ex]t10_d.south)
   \Edge([yshift=0.7ex]t11_e.north)([yshift=-0.6ex]t11_d.south)
   
   \Edge([yshift=-0.4ex]sos_1_d.north)([yshift=-0.6ex]sos_1_end.south)
   \Edge([yshift=-0.4ex]sos_2_d.north)([yshift=-0.6ex]sos_2_end.south)
   \Edge([yshift=-0.4ex]sos_dots_d.north)([yshift=-0.6ex]sos_dots_end.south)
   \Edge([yshift=-0.4ex]sos_n_d.north)([yshift=-0.6ex]sos_n_end.south)
   
   \node[draw=none, rectangle, rounded corners=0.0cm, minimum width=0.1cm, minimum height = 0.5cm, align=center](log_p) at (2, 1.25){\footnotesize $\log\left(P(\text{``positive''}\vert \overleftarrow{\text{input}})\right)$};
   \node[draw=none, rectangle, rounded corners=0.0cm, minimum width=0.1cm, minimum height = 0.5cm, align=center](dummy) at (-1, 1.0){{\color{vectorred} \small $\nabla$\textbf{ Update}}};
   \node[draw=none, rectangle, rounded corners=0.0cm, minimum width=0.1cm, minimum height = 0.5cm, align=center](log_p_dummy) at (-4.5, 1.25){};
   \Edge([yshift=0.3ex]t8_d.north)([yshift=1.0ex]log_p.south)
   \SetUpEdge[style={-, dashed}, lw = 0.75 pt, color = black,]
   \Edge([xshift=-1ex]log_p.west)(log_p_dummy.center)
   \Edge[style={bend right,out=-30,in=-135, dashed, ->}, color=black](log_p_dummy.center)([yshift=0.2ex]E2.north)
\end{tikzpicture}
}
\vskip -2ex
\caption{Illustration of the prompt-tuning approach used for parameter efficient fine-tuning of the models. The prompt tokens, depicted with orange hatching, are initialized as the beginning-of-sequence token embedding. These embeddings are subsequently perturbed by adding learned prompt embeddings. All weights are frozen during back-propagation except for the prompt embedding layer.}
\label{prompt_tuning_figure}
\end{figure*}
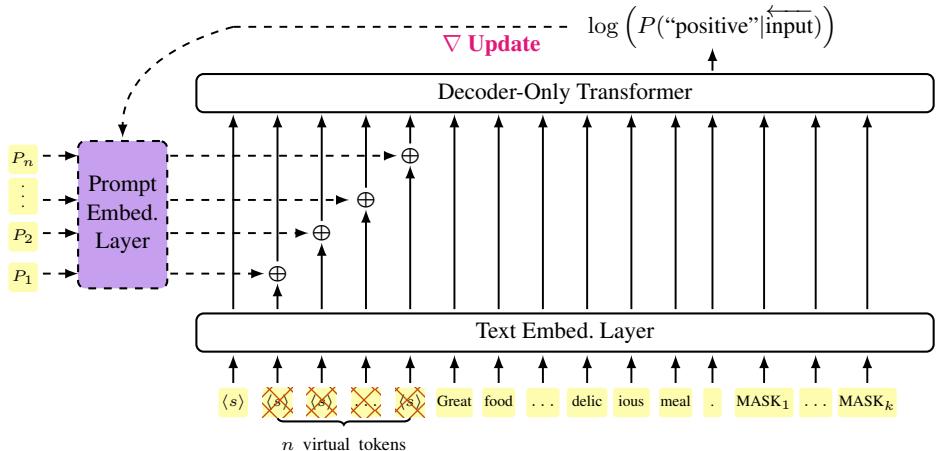

\subsection{Soft-prompt Tuning Details} 

The soft-prompting approach adds a series of tokens with trainable embeddings, $T = \{t_1,t_2,\ldots, t_n\}$, to the model input text $X$. Given a target token or set of tokens $Y$, the objective is to maximize the log-likelihood of the generation probability of $Y$ conditioned on the tokens, $T$, and input text, $X$, expressed as $P\left(Y \vert T; X\right)$. For the sentiment tasks examined here, the target tokens are \emph{positive}, \emph{negative}, and \emph{neutral}. An illustration of the prompt-tuning procedure is shown in Figure \ref{prompt_tuning_figure}. The weights of the underlying LM are frozen throughout the training process. Thus, producing task-specific representations does not explicitly modify biases inherited from the LM pre-training data. We hypothesize that when compared with full-model fine-tuning, this approach ensures a more accurate assessment of the bias innate to the LM. On the other hand, the optimized prompt embeddings help ensure that the model performs the downstream task as well as a fully fine-tuned model, which naturally reflects the settings of practical deployment. For training, a standard AdamW optimizer is used \cite{Loshchilov1}. We leveraged the JAX ML framework \cite{jax2018github} to achieve efficient model parallelism on TPUv3-8 devices and on machines with up to four A40 48GB GPUs.

As shown in Figure \ref{prompt_tuning_figure}, beginning-of-sequence tokens are used to provide initial embeddings for the continuous prompts. Each embedding is then additively perturbed by the trainable prompt embedding layer before flowing through the LM as usual, along with the remaining unmodified input-text tokens. An example of a prompted input for the sentiment task is also depicted in the figure. Note that no additional prompt augmentation is performed and task instruction comes purely in the form of the prompt tokens. Based on hyperparameter search results, the number of prompt tokens is fixed at $8$ for all experiments. Each prompt token is a dense vector with the same dimensionality as the embedding-space of the corresponding LM, which ranges from $1024$ to $5120$, depending on model size. Overall, the parameters learned are on the scale of $0.003\%$ of the full LM model weights.

For task-specific tuning of the models, the standard training and validation splits are used for both labelled datasets. The learning rate is optimized using validation accuracy. A concrete description of the hyperparameter sweep, along with the final parameters chosen appears in Appendix \ref{hyperparameter_details}. Given the inherent instability of prompt tuning, after hyperparameter selection, we tuned 15 different prompts, each with a different random seed. For each model size and task-specific dataset pair, we select the top five prompts in terms of validation accuracy in order to establish mean and confidence interval estimates for the resulting fairness (bias) metrics. Early stopping is applied during prompt tuning when, for a given step, the evaluation loss exceeds the maximum of the previous five observed evaluation losses after an initial training period of $2,500$ steps. All prompts are trained until the early stopping criterion is met.

\begin{figure*}[ht!]
    \centering
     \includegraphics[width=0.445\textwidth]{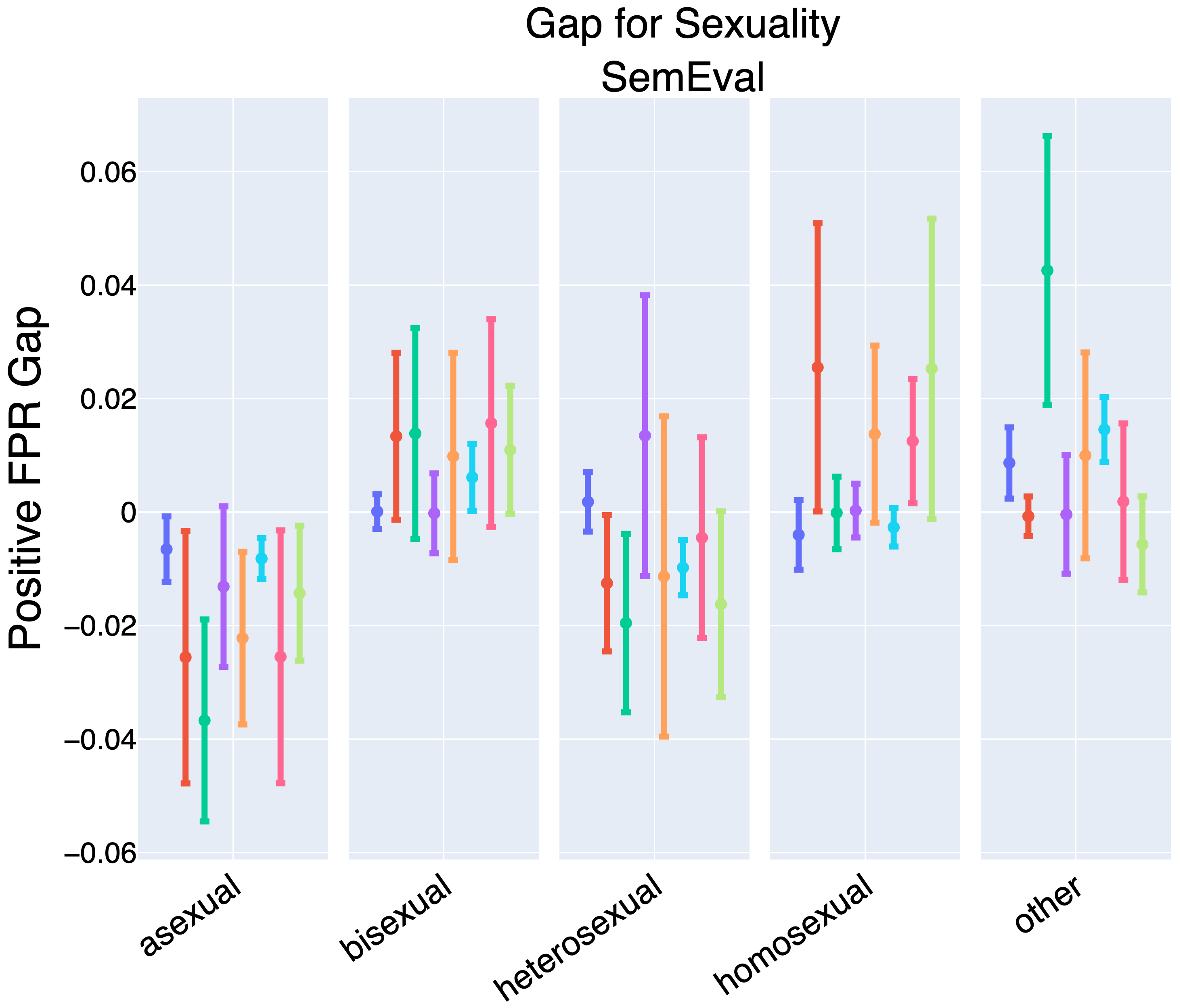}
    \includegraphics[width=0.54\textwidth]{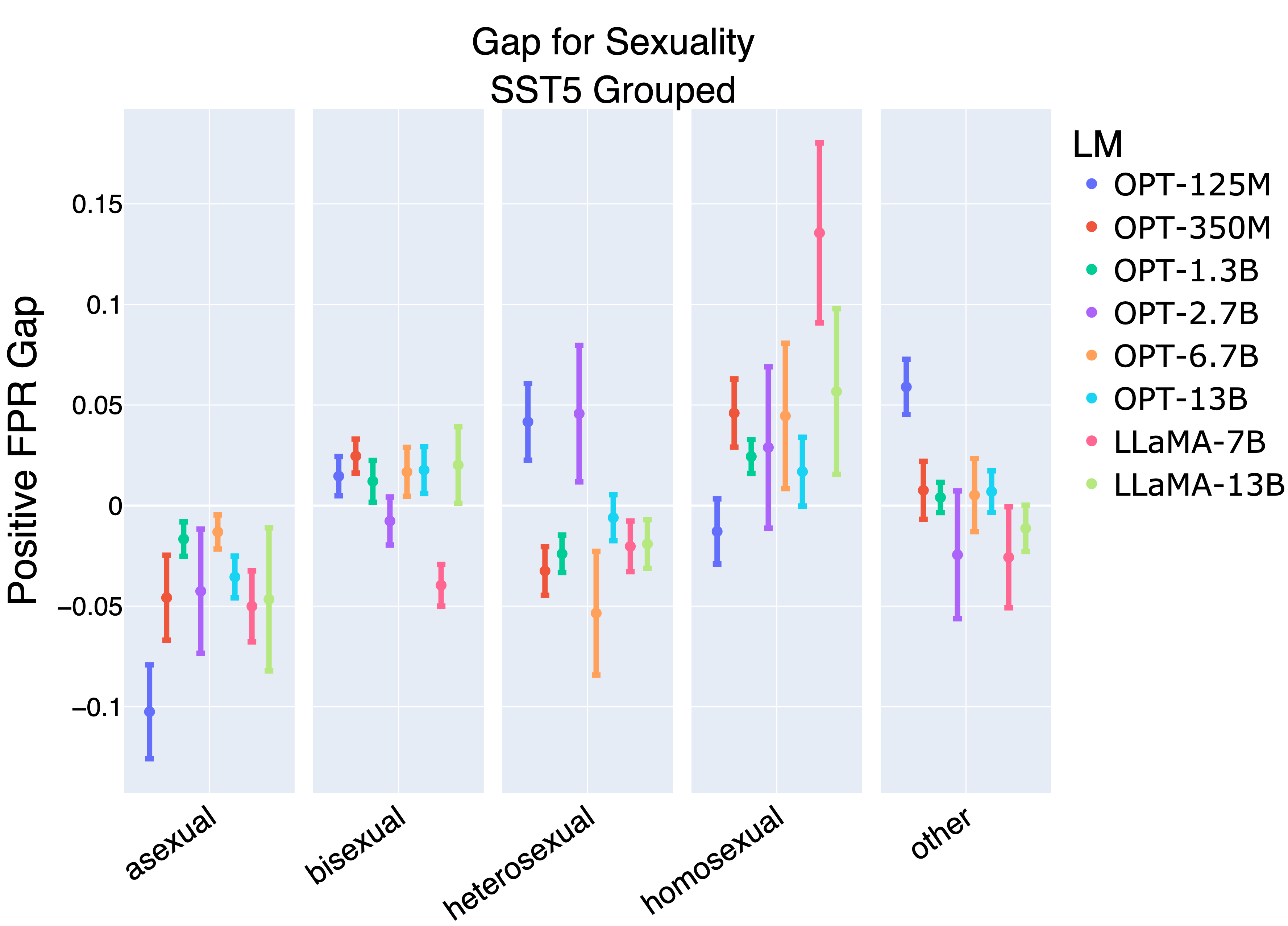}
    \caption{Positive FPR gap for the sensitive attribute of sexuality. Markers indicate average gap and bars are $95$\% confidence intervals. A positive gap indicates model errors that favor a group over others. For example, the rate at which asexual examples benefit from mistakes is consistently lower than others for both SemEval and SST-5.}
    \label{fig:sexuality_positive_FPR}
\end{figure*}

\begin{figure*}[ht!]
    \centering
     \includegraphics[width=0.445\textwidth]{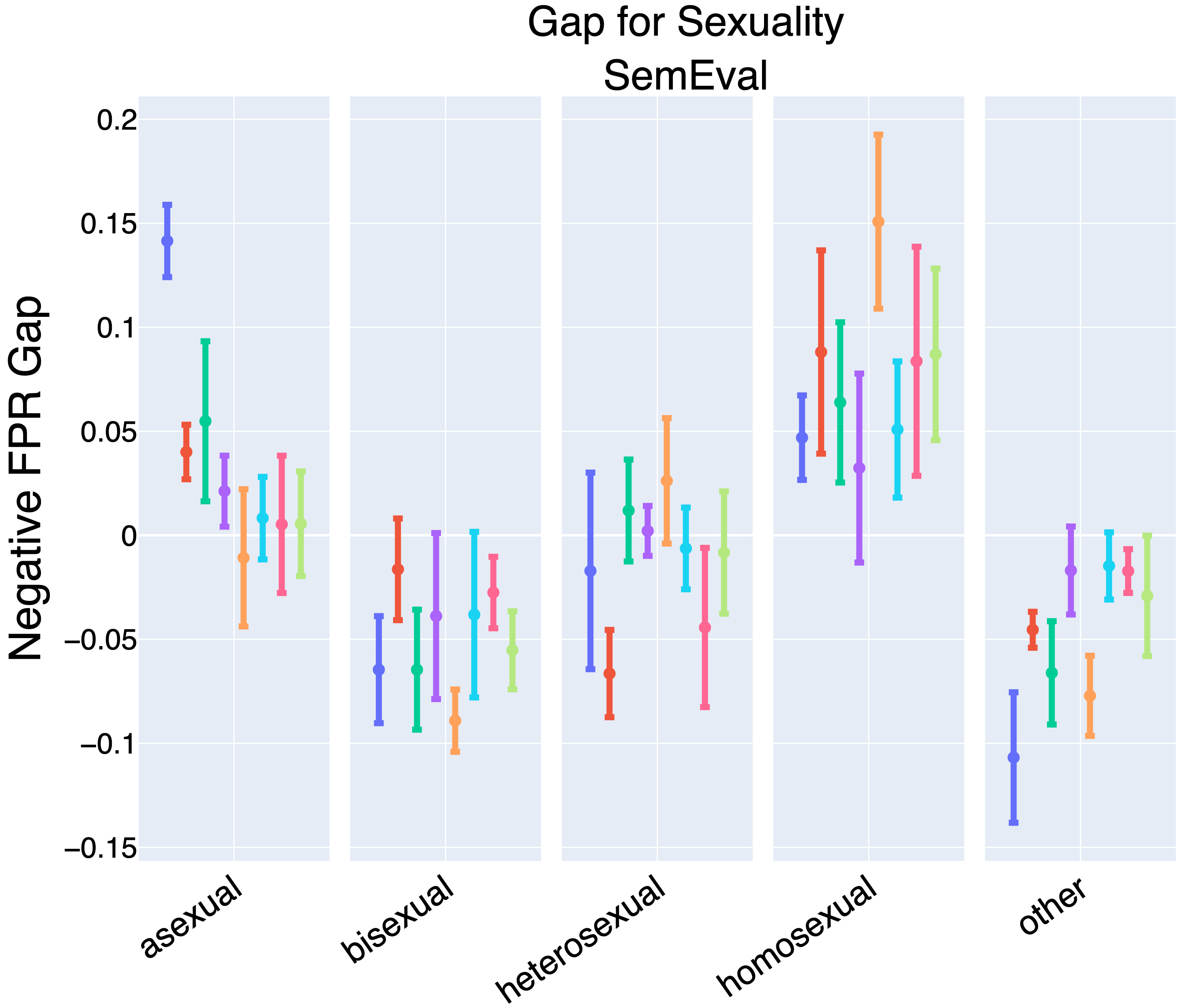} 
    \includegraphics[width=0.54\textwidth]{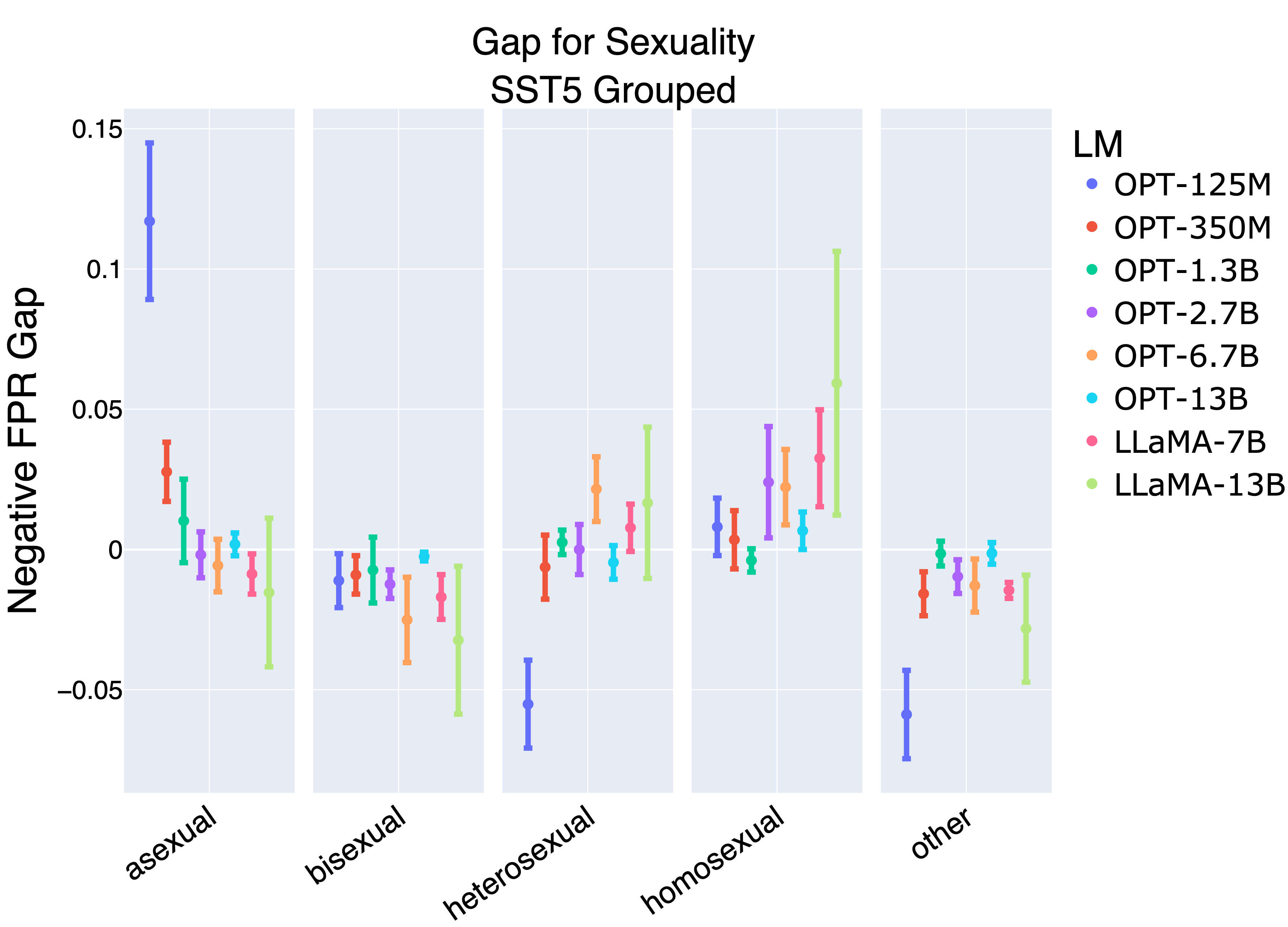}
    \caption{Negative FPR gap for the sensitive attribute sexuality. Markers indicate the average gap and bars are $95$\% confidence intervals. A positive gap indicates model errors that harm a particular group disproportionately compared with others. Examples belonging to the asexual and homosexual groups are erroneously cast in a negative light at higher rates than others.}
    \label{fig:sexuality_negative_FPR}
\end{figure*}

\section{Results} \label{results}

In this section, results are presented for different sensitive attributes by showing the FPR gap for the various protected groups when using the SemEval and SST-5 datasets for prompt tuning. We also consider the impact of tuning various model sizes of OPT and LLaMA on the metrics.

\subsection{Sexuality FPR Gaps} \label{sexuality_results}

In Figure \ref{fig:sexuality_positive_FPR}, the FPR gap for positive sentiment is shown for sexuality. Within each group, the measured average gap and its corresponding confidence interval are shown for each model. As discussed above, the \textit{Positive FPR Gap} measures the rate at which the model erroneously classifies negative or neutral statements associated with the protected group in a favourable light. Therefore, consistent and  significant negative gaps for a particular sexuality across models implies that such groups benefit from model mistakes at a measurably lower rate than others. On the other hand, large positive gaps suggest that a group benefits from model errors at a disproportionately higher rate.

Figure \ref{fig:sexuality_positive_FPR} shows that the rate at which examples belonging to the \textit{asexual} group benefit from model mistakes is consistently lower for models trained on both the SemEval and SST-5 datasets and across all model sizes. Somewhat surprisingly, in this measure, there is evidence to suggest that \textit{heterosexual} examples constitute an unfavoured group and do not benefit from model mistakes. However, the pattern is fairly weak. It is also interesting to note that examples from the \textit{bisexual} group benefit disproportionately from model mistakes in both datasets. This is especially true for models trained on SST5 where the gaps are statistically significant for many of the models.

The results in Figure \ref{fig:sexuality_negative_FPR} display the \textit{Negative FPR Gap}. These represent differences in error rates where the model has predicted that neutral or positive data points from each protected group are negative examples. Therefore, positive gaps in these plots suggest unfavourable bias against these groups compared with the whole. For smaller models it is evident that, as in Figure \ref{fig:sexuality_positive_FPR}, the \textit{asexual} group suffers from an elevated harmful error rate. Furthermore, examples from \textit{homosexual} group experiences large and statistically significant elevation in Negative FPR for both datasets considered and nearly all models. Two protected groups, \textit{bisexual} and \textit{other}, experience statistically significant decreases in the FPR measure for nearly all models across both datasets, markedly separating from other groups.

Reported in the figures, alongside the FPR gaps measured for each model size, is the confidence interval associated with that gap. For each group, Table \ref{tab:sexuality_stat_sig_gap_counts} displays the net number of times the gap was below or above zero, at $95$\% confidence. That is, for each significant gap below zero we subtract one, while one is added for significant gaps above zero. Values colored in red indicate the direction of the significant gaps that are considered harmful, while those in green denote potentially favourable treatment by the models, though this depends on how model results are used in practice. 

\begin{table}[ht!]
    \centering
    \resizebox{\columnwidth}{!}{\begin{tabular}{|c|c|c|c|c|}
        \hlinewd{1.3pt}
       Metric  & \multicolumn{2}{c|}{Positive FPR Gap} & \multicolumn{2}{c|}{Negative FPR Gap} \\
       \hlinewd{1.3pt}
       Group\textbackslash Dataset  & SemEval & SST-5 & SemEval & SST-5 \\
       \hline
       Asexual & {\color{red}-7} & {\color{red}-8} & {\color{red}4} & {\color{red}1} \\
       \hline
       Bisexual & {\color{darkgreen}1} & {\color{darkgreen}5} & {\color{darkgreen}-5} & {\color{darkgreen}-7} \\
       \hline
        Heterosexual & {\color{red}-3} & {\color{red}-3} & {\color{darkgreen}-2} & {0} \\
        \hline
       Homosexual & {\color{darkgreen}2} & {\color{darkgreen}5} & {\color{red}7} & {\color{red}5} \\
       \hline
       Other & {\color{darkgreen}3} & {0} & {\color{darkgreen}-6} & {\color{darkgreen}-6} \\
       \hline
        \hline
       Adult & {\color{darkgreen}1} & {0} & {\color{darkgreen}-6} & {\color{darkgreen} -2} \\
       \hline
       Old & {\color{red}-3} & {\color{red}-2} & {\color{red}2} & {\color{darkgreen}-1} \\
       \hline
       Young & {0} & {\color{darkgreen}1} & {\color{red}2} & {\color{red}3} \\
       \hline
    \end{tabular}}
    \caption{Net number of models (out of $8$) where the gaps for each protected group differ from zero at the $95$\% confidence level. Negative values imply that the gap was consistently below zero. Red numbers indicate that the direction of the gaps are harmful. The top five rows correspond to the sensitive attribute sexuality, while the bottom three are associated with age.}
    \label{tab:sexuality_stat_sig_gap_counts}
\end{table}

\begin{figure*}[ht!]
    \centering
     \includegraphics[width=0.445\linewidth]{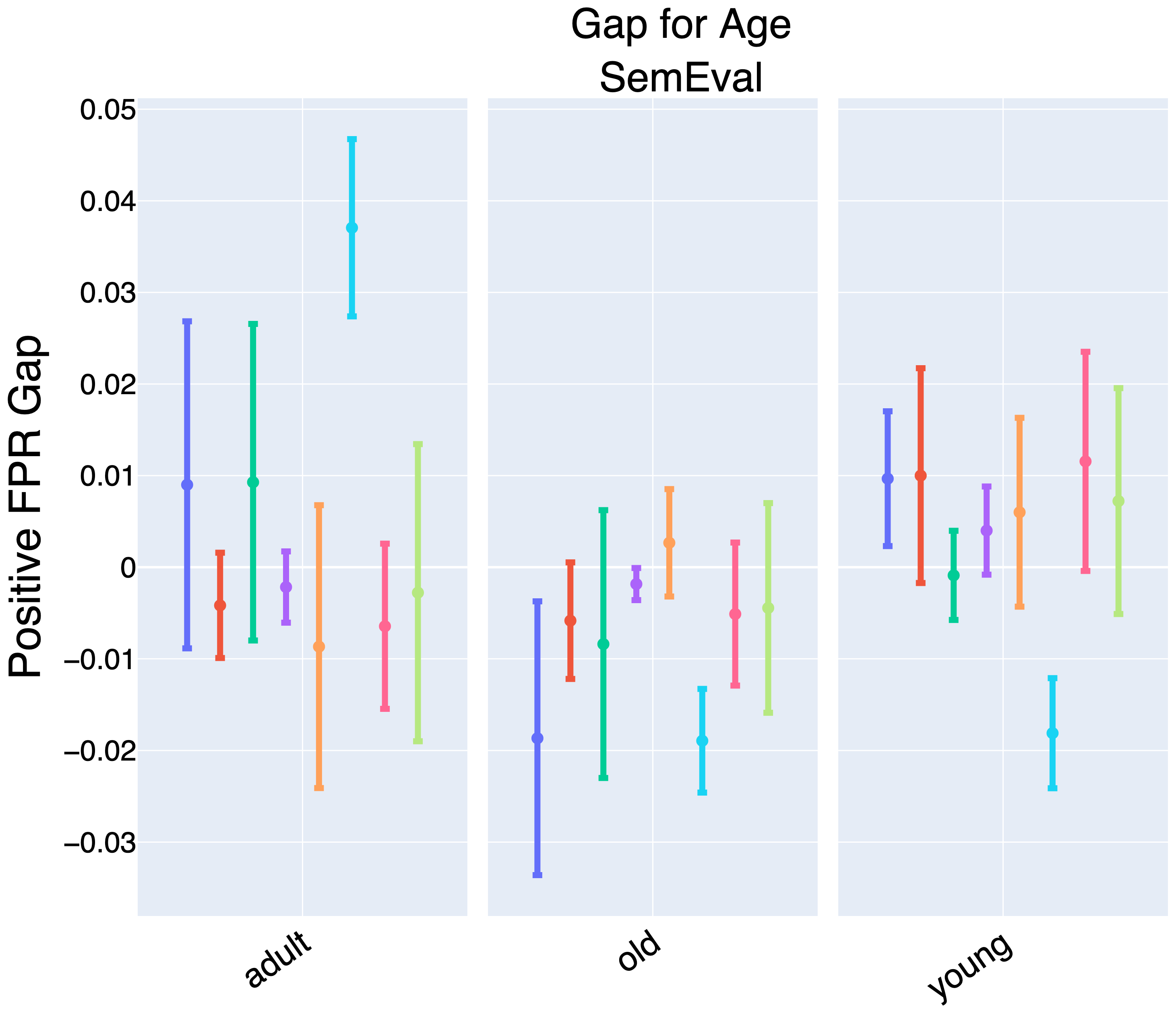}
    \includegraphics[width=0.54\textwidth]{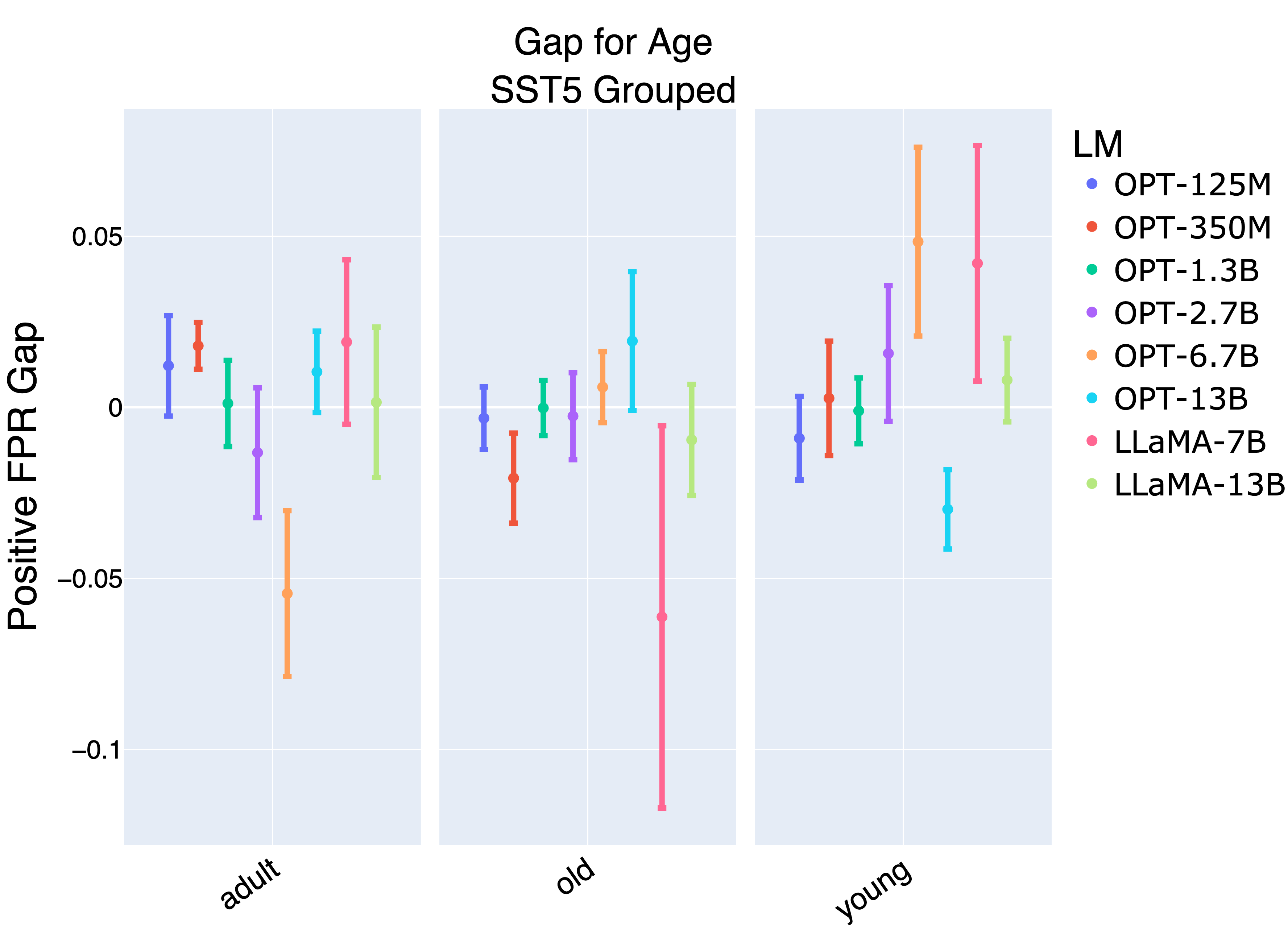}
    \caption{Positive FPR gap for the sensitive attribute of age. Markers indicate the average gap and bars are $95$\% confidence intervals. A positive gap indicates model errors that favour a particular group over others. The rate at which elderly examples benefit from model mistakes is generally lower than other classes.}
    \label{fig:age_positive_FPR}
\end{figure*}

\begin{figure*}[ht!]
    \centering
     \includegraphics[width=0.445\textwidth]{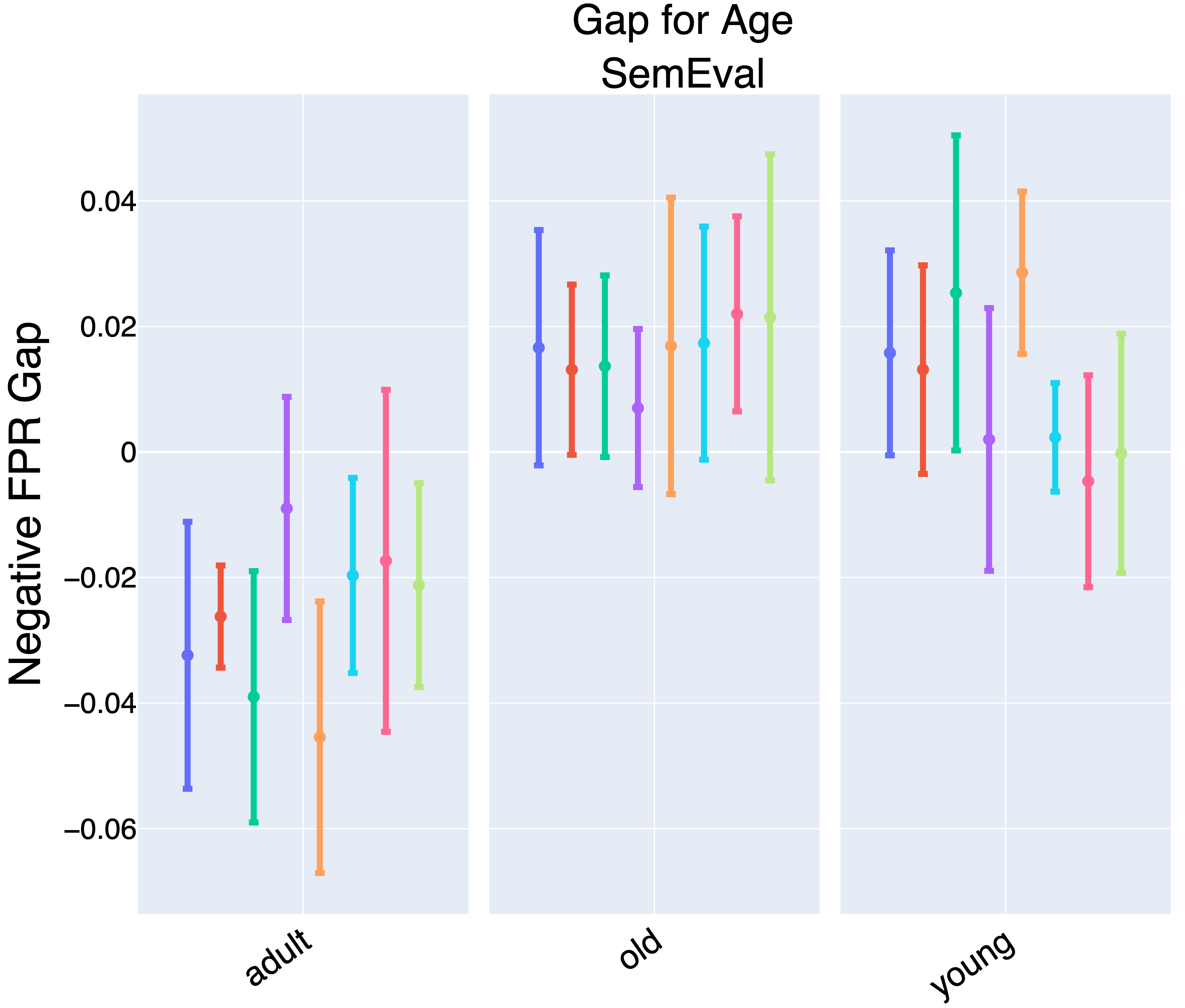}
    \includegraphics[width=0.54\textwidth]{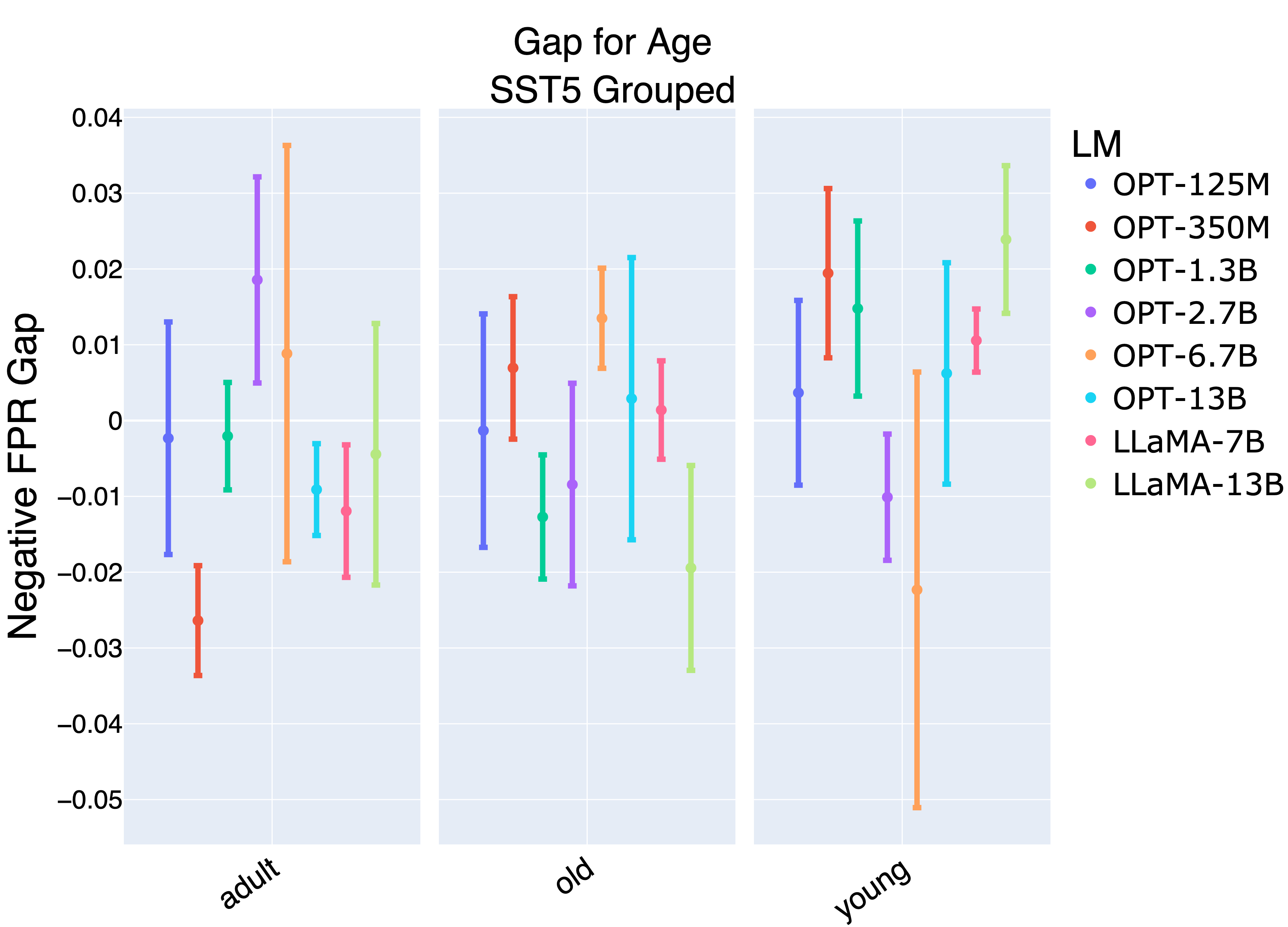}
    \caption{Negative FPR gap for the sensitive attribute of age. Markers indicate average gap and bars are $95$\% confidence intervals. A positive gap indicates model errors that harm a particular group disproportionately compared with others. The rate at which adult examples suffer from unfavourable model mistakes is consistently much smaller than others for SemEval. This conclusion is not as clear for SST-5.}
    \label{fig:age_negative_FPR}
\end{figure*}

For the \textit{asexual} and \textit{homosexual} protected groups, the experimental results strongly indicate potential harmful bias in the Positive FPR and Negative FPR Gaps, respectively. This is consistent across datasets, model sizes, and model types. On the other hand, the protected groups of \textit{bisexual} and \textit{other} consistently benefit from model mistakes at statistically significant, elevated rates in both gap measures for all experimental configurations.

Overall, in the experiments above, the observed gaps in FPR for both positive and negative classes are consistent across model type, model size, and datasets showing that prompt-tuning, as a fairness probe, is effective in revealing consistent inherited bias. Moreover, a number of protected groups experience statistically significant FPR gaps across all or nearly all of the different experimental setups.

\subsection{Age FPR Gaps}

As with the sensitive attribute of sexuality, the FPR Gaps for protected groups belonging to the age attribute are analyzed in this section. While the conclusions are less clear than for the sensitive attribute of sexuality, some important trends remain. Figure \ref{fig:age_positive_FPR} shows the FPR Gap measured for the positive class. When considering results from the SemEval dataset, a marked decrease in FPR is present for the \textit{old} group of examples. This trend is also present for the SST-5 dataset, though it is weaker. On the other hand, when considering the measurements in Figure \ref{fig:age_negative_FPR}, the \textit{adult} group is impacted by errors casting them in a negative light at a significantly lower rate than the other groups for the SemEval dataset. In addition, the \textit{old} and \textit{young} groups generally suffer from an elevated probability of such errors, though the gaps are not always statistically significant when confidence intervals are considered. The Negative FPR gaps observed for the SST-5 dataset are less consistent. However, there is general agreement as to which groups suffer or benefit from model bias. That is, examples from the \textit{adult} group are favoured and those from the \textit{young} group receive unfavourable errors, though the way in which the bias is manifested is slightly different depending on the underlying prompt-tuning dataset. Table \ref{tab:sexuality_stat_sig_gap_counts} reinforces this conclusion. Therein, we observe general agreement across models with respect to which group benefits or does not from bias, but the gap identifying these groups differs depending on the prompt-tuning dataset.

In Appendix \ref{disability_plots_appendix}, additional FPR gap results are presented for the sensitive attribute of disability. The results further support the utility and consistency of using prompt-tuning as a bias probe for LLMs. The measured gaps are largely consistent across model type and size, as well as tuning dataset, within groups. Furthermore, many of the measured gaps are statistically significant.

\section{Conclusions and Discussion} \label{conclusion}

In this paper, we have demonstrated the benefits of leveraging soft-prompt tuning as a mechanism for bias quantification in LLMs. The method offers several advantages over manual prompt optimization including removing the need for prompt design, better task performance, and limited injection of external bias. Moreover, it is faster and more efficient than full-model fine-tuning, with equivalent or better performance. Thus, uncovered biases more accurately reflect real-word deployment. 

The results show that, for example, within the sensitive attributes of sexuality, age, protected groups under the terms \textit{asexual}, 
\textit{homosexual}, and old receive unfavourable treatment, compared with other groups, consistently across datasets, model size, and model type. However, the following points should be considered for a comprehensive analysis.

\subsection{Multidimensional Aspects of the Experiments}

While in this paper, we have explored the utility of a state-of-the-art soft-prompt tuning technique, the chosen downstream task is, in itself, challenging yet impactful. This coupling makes the exploration interesting but the analysis of the results is multidimensional across datasets, templates, prompt-tuning choices, sensitive attributes, their protected groups, models, fairness (bias) metrics, and their graphical representations. We have tried our best to present the results in the most comprehensive way.

\subsection{Template Design}

Throughout the analysis, we use the fairness probing templates of \cite{Czarnowska1}. They provide an important baseline for the experiments, but consist of simple sentences, which are often easily understood by the LLMs. In spite of this, consistent and significant disparities are observed for certain groups. However, this may be the cause of less conclusive results for some groups. In future work, we aim to perform experiments using more complicated templates.

\subsection{Types of Biases} Many papers \cite{Czarnowska1} rely on absolute values of the metric disparities to simply reveal the presence and potential magnitude of bias. We use a directional bias measure to identify the favoured and unfavoured groups, providing more precise bias analysis of the LLMs. It should be mentioned that a group that is flagged as a favourable group may be flagged as unfavourable by using a different bias quantification metric or considering a different downstream task. Thus, different bias quantification formulations \cite{kalantari2021Underdiagnosis} might not be concurrently achievable.

\subsection{Impact of Soft-prompt Tuning on Bias}

Fairness evaluation through prompting, and prompt tuning in particular, offers several advantages over traditional fine-tuning approaches. Foremost among them is that it is significantly more resource efficient while producing comparable downstream task performance \cite{Lester1} in large models. In addition, continuous prompt tuning minimizes the potential influence of biases existing in the supervised training tasks by restricting the number of learned parameters. Finally, it removes the human element of prompt design, eliminating another avenue for bias introduction outside of the LLM itself. However, it should be noted that we performed soft-prompt tuning on standard datasets that were generated from tweets (SemEval) and movie reviews (SST-5). The quality of these datasets has a strong impact on the soft-prompts produced. It is worth exploring how a better quality dataset (if available) impacts the performance of the downstream task and the biases observed. 

In addition to the directions mentioned above, we plan to extend our work by including a broader range of LMs, expanding to more sensitive attributes, considering more bias metrics, and incorporating other downstream tasks. This is an effort to make the use of LLMs safer and more ethical in real-world deployment.

\bibliographystyle{acl_natbib}
\bibliography{custom}

\begin{figure*}[hb!]
    \centering
     \includegraphics[scale=0.09]{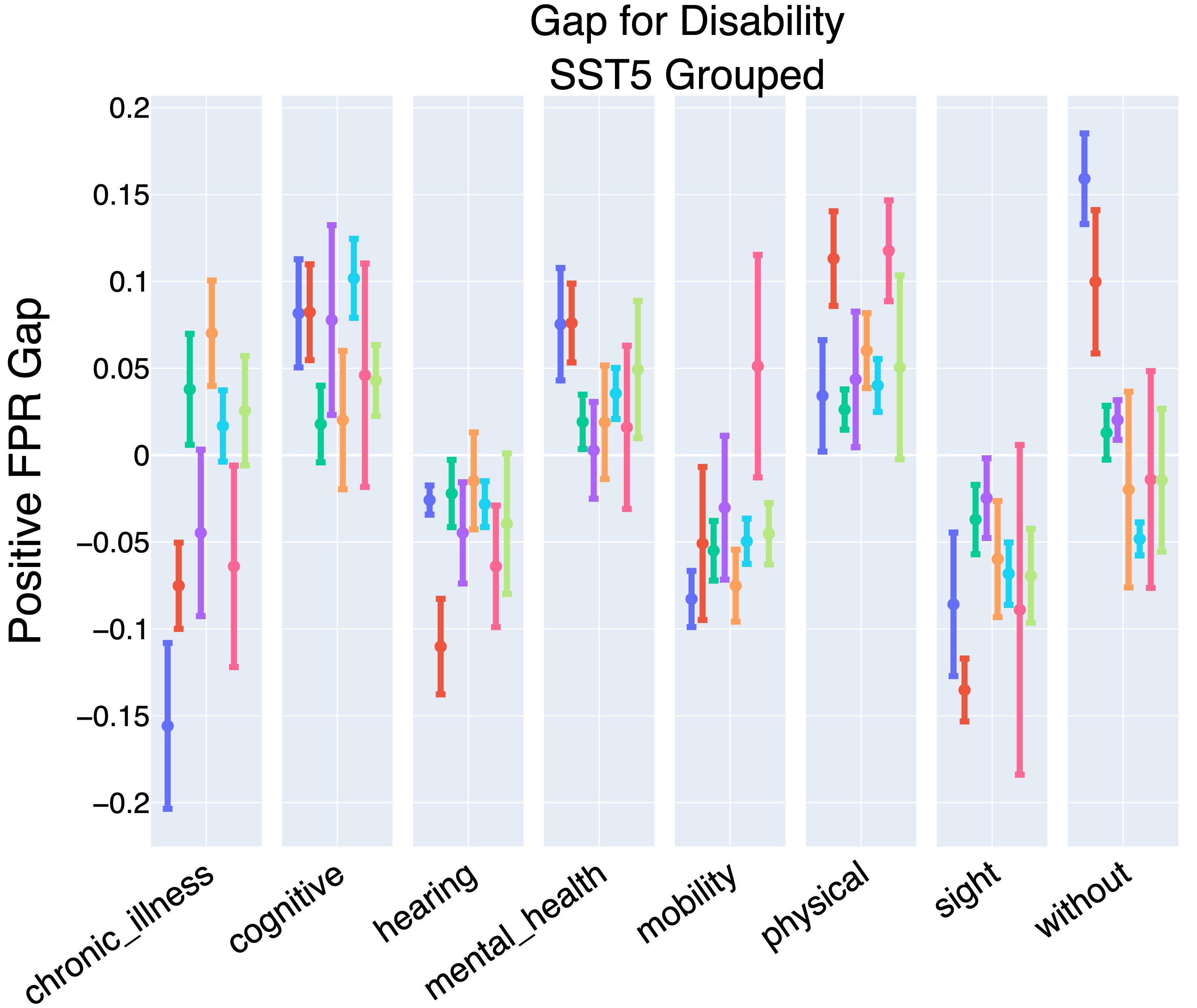}
    \caption{Positive FPR gap for the sensitive attribute of disability. Markers indicate average gap and bars are $95$\% confidence intervals.}
    \label{fig:disability_sst5_positive_FPR}
\end{figure*}

\appendix

\section{Fairness Vocabulary} \label{fairness_metrics_def}
\textbf{Sensitive attribute}: An attribute within which social biases may be exhibited. Examples of sensitive attributes include age, disability, gender, nationality, race, religion, and sexuality.\\ 
\textbf{Protected group}: Each sensitive attribute consists of different protected groups over which model behaviour should remain consistent. For example, the sensitive attribute \textit{age} may consist of \{\textit{child, teenager, adult, older adult}\}.

\section{Hyperparmeter Details} \label{hyperparameter_details}
We conducted a hyperparameter search over the validation split of SemEval and SST5 for the following possible learning rate values: 0.01, 0.001, 0.0001. We list the hyperparameters selected for each model in Table \ref{tab:hyperparameters}. The number of prompt tokens for all models is fixed at $8$. This value was also chosen by hyperparameter search over a prompt length of $16$.

\begin{table}[ht!]
    \centering
    \begin{tabular}{c|c}
        \hlinewd{1.3pt}
         Model & LR \\
         \hlinewd{1.3pt}
         OPT-125M & 0.001 \\
         \hline
         OPT-350M & 0.001 \\
         \hline
         OPT-1.3B & 0.001 \\
         \hline
         OPT-2.7B & 0.001 \\
         \hline
         OPT-6.7B & 0.001 \\
         \hline
         OPT-13B & 0.0001 \\
         \hlinewd{1.3pt}
         LLaMA-6.7B & 0.0001 \\
         \hline 
         LLaMA-13B & 0.0001 \\
         \hline
    \end{tabular}
    \caption{Learning rate applied for each model during prompt tuning.}
    \label{tab:hyperparameters}
\end{table}

\section{Datasets} \label{dataset_description}

For each model, we tune continuous prompts on two sentiment datasets. These are the SemEval-2018 1-Valence Ordinal Classification Task \cite{SemEval2018Task1} (SemEval) and Stanford Sentiment Treebank Five-way dataset (SST-5)\cite{socher-etal-2013-recursive}. The SemEval dataset is a collection of English tweets with integer labels ranging from $[-3, 3]$. Following \cite{Czarnowska1}, these labels are condensed by the mapping \{\textit{Negative} 0: [-3, -2], \textit{Neutral} 1: [-1, 0, 1], \textit{Positive} 2: [2, 3]\}. The labels of SST-5 (\textit{very positive}, \textit{positive}, \textit{neutral}, \textit{negative}, \textit{very negative}) are based on brief English movie reviews and, therefore, constitute a very different underlying corpus. As with the SemEval valence labels, the five-way annotations of SST-5 are collapsed to three-way classification by retaining the \emph{neutral} label and mapping positive and negative polarity of any kind simply to \emph{positive} or \emph{negative} classes, respectively.

\section{Gap Results for Disability} \label{disability_plots_appendix}

\begin{figure*}[ht!]
    \centering
    \includegraphics[scale=0.09]{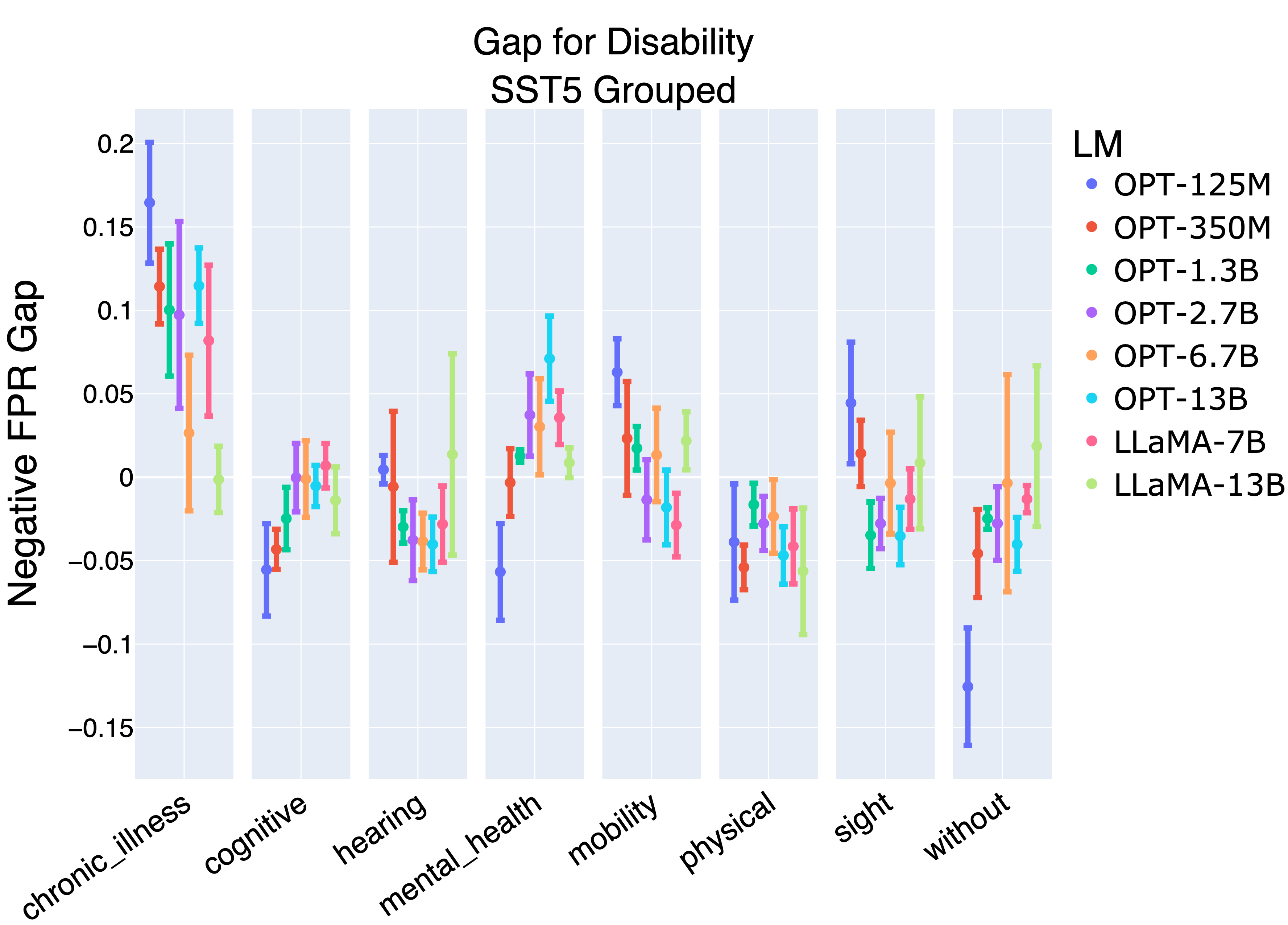}
    \caption{Negative FPR gap for the sensitive attribute of disability. Markers indicate average gap and bars are $95$\% confidence intervals.}
    \label{fig:disability_sst5_negative_FPR}
\end{figure*}

In this section, the protected groups belonging to the sensitive attribute of disability are considered. Figures \ref{fig:disability_sst5_positive_FPR} and \ref{fig:disability_sst5_negative_FPR} and display the measured positive and negative FPR gaps, respectively, for OPT and LLaMA models prompt-tuned on the SST-5 dataset. In terms of positive FPR, there are many statistically significant negative gaps for examples associated with \textit{hearing}, \textit{mobility}, and \textit{sight}  impairment. Alternatively, positive gaps are seen for the groups denoted by \textit{cognitive} and \textit{physical} disabilities. 

For negative FPR, a large positive gap is seen for examples belonging to the group \textit{chronic\_illness}. Small, but statistically significant negative, gaps for \textit{hearing} and \textit{physical} impairments are present across the various experimental configurations.

\end{document}